\documentclass[11pt,a4paper]{article}
\usepackage[hyperref]{emnlp2020}
\usepackage{times}
\usepackage{latexsym}

\usepackage{booktabs}
\usepackage{amsmath}
\usepackage{amsfonts}
\usepackage{microtype}
\usepackage{graphicx}

\aclfinalcopy 

\newcommand*{\affmark}[1][*]{\textsuperscript{#1}}
\newcommand*{\affaddr}[1]{#1}

\title{Adapting Event Extractors to Medical Data: Bridging the Covariate Shift}

\author{Aakanksha Naik\affmark[1] \And Jill Lehman\affmark[2] \\ 
  \affaddr{\affmark[1]Language Technologies Institute, Carnegie Mellon University} \\
  \affaddr{\affmark[2]Human-Computer Interaction Institute, Carnegie Mellon University} \\
  \texttt{\{anaik,jfl,cprose\}@andrew.cmu.edu} \And Carolyn Ros\'e\affmark[1]}

\date{}

\begin{document}
\maketitle
\begin{abstract}
We tackle the task of adapting event extractors to new domains without labeled data, by aligning the marginal distributions of source and target domains. As a testbed, we create two new event extraction datasets using English texts from two medical domains: (i) clinical notes, and (ii) doctor-patient conversations. We test the efficacy of three marginal alignment techniques: (i) adversarial domain adaptation (ADA), (ii) domain adaptive fine-tuning (DAFT), and (iii) a novel instance weighting technique based on language model likelihood scores (LIW). LIW and DAFT improve over a no-transfer BERT baseline on both domains, but ADA only improves on clinical notes. Deeper analysis of performance under different types of shifts (e.g., lexical shift, semantic shift) reveals interesting variations among models. Our best-performing models reach F1 scores of 70.0 and 72.9 on notes and conversations respectively, using \emph{no} labeled data from target domains.
\end{abstract}
\section{Introduction}
Events are an important phenomenon in the field of computational semantics. They offer an intuitive mechanism for constructing structured representations of text, which can be used for downstream tasks such as question answering and summarization. Events also embody a crucial function of language: the ability to report happenings. Narratives from many diverse domains (e.g., news articles, literary texts, clinical notes) use events as basic building blocks. These characteristics make event extraction a key sub-task of interest for text understanding pipelines in multiple domains. Despite its importance, building high-performing and generalizable systems for event extraction has remained an elusive goal. One of the major hurdles is that the notion of what counts as an \emph{important event} is usually task-specific or domain-specific (sometimes both). For example, to build a system that can track a patient's disease progression from clinical notes, event extractors only need to focus on extracting medical events relevant to that illness. This task/domain specificity has encouraged prior work to focus on specific event types \cite{grishman-sundheim-1996-message,doddington2004automatic,kim2008corpus} or domains \cite{pustejovsky2003timebank,sims-etal-2019-literary}. Owing to this narrow focus, supervised event extractors often fail to adapt to new domains or event types \cite{keith-etal-2017-identifying}. Unsupervised event extractors that use syntactic rule-based modules \cite{sauri-etal-2005-evita,chambers-etal-2014-dense}, conversely, have a tendency to over-generate by labeling most verbs and nouns as events. 

In this work, we try to achieve a balance between these extremes by building \emph{adaptable} event extractors using unsupervised domain adaptation techniques. We also study the behavior of these techniques under various types of linguistic shifts (e.g., lexical shift, semantic shift) to gain insight into differences between them. Exploring adaptability under low-supervision is crucial, since sourcing annotated data for new domains, especially medical texts, can be expensive and time-consuming. We formulate event extraction as the task of labeling \textbf{triggers}, i.e., words which instantiate an event. For example, in the sentence ``She was diagnosed with cancer,'' \emph{diagnosed} and \emph{cancer} are triggers, referring to ``diagnosis'' and ``illness'' events respectively. Throughout our work, we model event trigger labeling as token-level classification. 

To test adaptability, we create new event extraction test sets using English texts from two diverse medical domains: (i) clinical notes, and (ii) doctor-patient conversations. We develop comprehensive event annotation guidelines, based on TimeML \cite{pustejovsky2003timeml} and Thyme-TimeML \cite{styler-iv-etal-2014-temporal} (\S\ref{sec:dataset}), and use them to annotate 45 documents from each domain. As a baseline, we train a BERT-based event extraction model on English news articles from TimeBank \cite{pustejovsky2003timebank}, which is labeled using TimeML, and test its performance on our datasets. To improve this out-of-domain baseline performance, we tackle the problem of covariate shift, i.e., differences between marginal distributions of source (news) and target domains (notes or conversations). We experiment with three marginal alignment techniques: (i) adversarial domain adaptation (ADA) \cite{ganin2015unsupervised}, (ii) domain-adaptive fine-tuning (DAFT) \cite{han-eisenstein-2019-unsupervised}, and (iii) a novel instance weighting scheme using language model likelihood scores (LIW). ADA is task-guided since it jointly performs alignment and task training. DAFT and LIW are task-agnostic, performing alignment and task training sequentially. 

Our results show that DAFT and LIW improve over BERT on both domains, whereas ADA only improves on clinical notes. Across domains, there is no clear winner, with ADA and DAFT performing best on notes and conversations respectively. Analyzing covariate shift at different levels (e.g., lexical shift, semantic shift), we uncover interesting patterns such as the ability of models to leverage sub-word morphology to generalize to some technical terms in clinical notes, and LIW's performance improvement on long-term state events (e.g., chronic illnesses). Our best models achieve F1 scores of 70.0 and 72.9 on notes and conversations respectively with \emph{no} training data. 

\section{Related Work}
\subsection{Event Extraction}
Most prior event extraction work has focused on news articles, resulting in the development of several datasets \cite{onyshkevych-etal-1993-tasks-domains,grishman-sundheim-1996-message,pustejovsky2003timebank,doddington2004automatic,lee-etal-2012-joint,cybulska-vossen-2014-using,mitamura2016overview}. Recently, event extraction has also been explored in other domains such as biology \cite{wattarujeekrit2004pasbio,kim2008corpus,kim-etal-2009-overview,berant-etal-2014-modeling}, Wikipedia articles \cite{araki-mitamura-2018-open}, social media data \cite{ritter2012open,li-etal-2014-major,jain-etal-2016-towards} and literary novels \cite{sims-etal-2019-literary}. Aside from data domain, event extraction paradigms (both datasets and tools) differ along three major axes: (i) event extraction granularity, (ii) event representation, and (iii) event categorization (ontology). We briefly describe these axes to contextualize our choice of event paradigm.

Event extraction granularity divides extraction paradigms into two types: (i) document-level paradigms which assume that a piece of text refers to a single event \cite{grishman-sundheim-1996-message}, and (ii) sentence-level paradigms which assume that a single sentence describes one or more events. Event representation divides extraction paradigms into two types: (i) span-based paradigms which represent events by marking text spans that refer to events, called \textbf{triggers} or \textbf{nuggets} \cite{pustejovsky2003timeml,mitamura-etal-2015-event,ogorman-etal-2016-richer}, and (ii) structured paradigms which represent events by marking text spans and adding additional arguments (e.g., participants, location etc.) to create a structured template \cite{grishman-sundheim-1996-message}. Event categorization divides extraction paradigms into two types: (i) ontology-driven paradigms that are limited to specific event types \cite{grishman-sundheim-1996-message,doddington2004automatic}, and (ii) ontology-free paradigms that do not place type restrictions \cite{pustejovsky2003timebank,araki-mitamura-2018-open}.  

We use a sentence-level, span-based, ontology-free event extraction paradigm. Sentence-level extraction suits our domains of interest since notes and conversations tend to discuss multiple events. Span-based and ontology-free extraction allows us to develop adaptable coding guidelines since event arguments and types are usually domain-specific or task-specific. This adaptability sets our work apart from other prior work on medical event extraction such as adverse drug event extraction \cite{nikfarjam2015pharmacovigilance,sarker2015portable,cocos2017deep,henry2018n2c2} and personal event extraction from online support groups \cite{wen-etal-2013-extracting,naik-etal-2017-extracting}, which focus on specific event types. Our guidelines draw heavily from the Thyme-TimeML guidelines \cite{styler-iv-etal-2014-temporal} used by the Clinical TempEval challenges on event ordering in clinical notes \cite{bethard-etal-2015-semeval,bethard-etal-2016-semeval,bethard-etal-2017-semeval},\footnote{We provide a detailed comparison with this work in \S\ref{ssec:annoguide}.} but also cover event extraction in a novel domain: doctor-patient conversations. 
\subsection{Unsupervised Domain Adaptation}
Unsupervised domain adaptation is the task of transferring a model from a source domain to a target domain, using only unlabeled data from the target domain, by aligning source and target distributions. Early approaches such as structural correspondence learning (SCL) \cite{blitzer-etal-2006-domain,blitzer-etal-2007-biographies} tried to solve this by mapping source and target examples into a shared \textbf{pivot feature} space, where pivot features are selected to be features that behave the same way for discriminative learning in both domains (e.g., sentiment terms such as \emph{amazing} and \emph{great} show similar behavior for sentiment analysis across domains). With advances in neural representation learning, autoencoder-based methods \cite{glorot2011domain,chen2014marginalized}, neural SCL \cite{ziser-reichart-2017-neural}, adversarial domain adaptation \cite{ganin2015unsupervised,ganin2016domain} and LM fine-tuning methods \cite{han-eisenstein-2019-unsupervised,gururangan2020dont} have shown success in learning a shared space in which source and target domains are aligned. We use adversarial domain adaptation (ADA) and domain adaptive fine-tuning (DAFT) because these methods have shown promise on sequence labeling tasks \cite{gui-etal-2017-part,han-eisenstein-2019-unsupervised,naik2020open}, and offer an interesting contrast between approaches that jointly perform alignment and task training (ADA) and approaches that perform these steps sequentially (DAFT).


\section{Dataset Creation}
\label{sec:dataset}
To test adaptability of event extraction models, we create a testbed using data from two domains:\\
\noindent
\textbf{1. Clinical Notes:} Clinical notes are records documenting physician observations from their interactions with patients. They usually detail various aspects of a patient's care such as present illness, symptoms, medical history, treatments, and test results. They share a thematic structure, though particular specialties (e.g., cardiology) and institutions often incorporate their own modifications. We collected a set of 4999 de-identified clinical notes from 40 specialties, by scraping mtsamples.\footnote{https://www.mtsamples.com/} Average length of a clinical note is 652 tokens.\\
\noindent
\textbf{2. Doctor-Patient Conversations:} This data contains human-transcribed, de-identified conversations recorded during physician-patient visits. The conversations often follow a similar schema, with patients describing their symptoms, doctors inquiring about ongoing treatments, and then suggesting potential follow-up treatments/tests. We use a proprietary dataset of 63,540 conversations covering 53 specialties. Average conversation transcript length is 2309 tokens.\\
These domains exhibit different types of linguistic shifts from the source (news). While both domains exhibit a shift in vocabulary, it is more pronounced in clinical notes since they are written by doctors (experts) who use highly technical terms. Conversely, shifts in syntax are more pronounced in conversations due to the prevalence of repetition, back-channeling, interruptions etc. Semantic shifts are more pronounced in conversations since they contain a higher proportion of hypothetical statements (e.g., when doctors ask questions, make requests or ``think out loud'') than both notes and news articles which tend to serve as records of actual events. To better evaluate model performance on linguistic shifts, we control for topical variation across domains by limiting our focus to 3 specialties: Cardiovascular/Pulmonary (Cardio), Obstetrics/Gynaecology (Obgyn) and Hematology/Oncology (Onco). These specialties are well-represented in both notes and conversations, and cover events with a variety of temporalities ranging from intervals with fixed duration (e.g., pregnancy), to intervals with indeterminable endpoints (e.g., long-term cardiac failure). Table~\ref{tab:specialty} gives an overview of the number of notes and conversations in each specialty. 

\begin{table}[]
    \centering
    \begin{tabular}{lll}
      \toprule \textbf{Specialty} & \textbf{\#Notes} & \textbf{\#Convos} \\ \midrule
        \textbf{Cardio} & 372 & 4876\\
        \textbf{Obgyn} & 160 & 1784\\
        \textbf{Onco} & 90 & 7177\\ \bottomrule
    \end{tabular}
    \caption{Domain-wise raw data statistics for chosen medical specialties}
    \label{tab:specialty}
\end{table}
\subsection{Developing Event Annotation Guidelines}
\label{ssec:annoguide}
We develop a set of coding guidelines for the task of annotating event triggers in documents from these two domains. Our coding guidelines build upon TimeML \cite{pustejovsky2003timeml}, a rich specification language for annotation of events and temporal expressions in text,\footnote{The complete TimeML coding manual is available here: \url{https://catalog.ldc.upenn.edu/docs/LDC2006T08/timeml_annguide_1.2.1.pdf}} and Thyme-TimeML \cite{styler-iv-etal-2014-temporal}, a variant of TimeML developed for clinical notes. We start with these guidelines because they use a syntax-driven domain-agnostic definition of events, allowing for an adaptable annotation scheme. In TimeML, the term \emph{event} refers to situations that \emph{happen} or \emph{occur}, or circumstances in which something \emph{obtains} or \emph{holds true}. This is a broad definition, consistent with Bach's definition of \textbf{eventualities} \cite{bach1986algebra}, and the idea of \textbf{fluents} \cite{mccarthy2002actions}. Events can be expressed in text by means of tensed or untensed verbs, nominalizations, adjectives, predicative clauses or prepositional phrases. TimeML describes rules to annotate events in all these syntactic categories. \newcite{styler-iv-etal-2014-temporal} adapted these rules for clinical notes. They focused on the THYME corpus of 1254 de-identified notes from the Mayo Clinic, representing two fields in oncology: brain cancer and colon cancer. As a first step, we annotate one document from each of our domains following TimeML and Thyme-TimeML rules. During this phase, we identify cases where it is reasonable to deviate from these guidelines. 

\noindent
\textbf{Deviations from TimeML:} Our guidelines\footnote{Our complete coding manual, including example annotations, is available here: \url{http://bitly.ws/9uxq}.} differ from TimeML in their treatment of two categories:\\
\noindent
\textbf{1. Activity patterns:} Activity patterns are events that are neither pure generics\footnote{Pure generics are events which discuss illnesses/treatments in general, and are not associated with a specific person and time. For example, ``there is a \emph{benefit} to systemic adjuvant \emph{chemotherapy}.''}, nor single events clearly positioned in time. For example, consider the sentence ``I \emph{take} my blood pressure regularly.'' The event \emph{take} is not grounded in time. It is also not a pure generic event as it is definitely associated with the speaker. Such events are \emph{not} annotated in TimeML. However, in our data, these activity patterns occur frequently in crucial contexts such as taking medications, following lifestyle changes suggested by doctors, measuring vital signs, etc.\\
\noindent
\textbf{2. Long-term states:} Because TimeML was geared towards the task of temporal ordering, it strictly restricted annotation of stative events to the following types: (i) states associated with a temporal expression, (ii) states undergoing a change within the document, (iii) states introduced by other events, since those can offer temporal cues, and (iv) states associated with the document creation time. However, many stative events in our data don't fit within these strict parameters, but are nevertheless important. The most crucial category is states associated with long-term ongoing illnesses (e.g., ``The patient has a long history of \emph{COPD}'').

These event categories are not specific to medical domains only. For example, long-term state events might be salient when extracting personal events from biographies.\footnote{e.g., ``Bill Gates is currently \emph{employed} full-time at the Bill and Melinda Gates Foundation.''} Similarly activity patterns might be salient when extracting events from scientific procedure manuals.\footnote{``\emph{Repeat} step 5 daily, over a period of 30 days.''} Considering these scenarios, we add rules to extract these two categories of events. We also expand syntactic rules to cover constructions unique to doctor-patient conversations such as repetition, especially for instructions, and hypothetical event annotation in utterances when doctors are ``thinking out loud''. 
\noindent \\
\textbf{Deviations from Thyme-TimeML:} Our guidelines differ from Thyme-TimeML in their treatment of two categories:\\
\noindent
\textbf{1. Generic events: }Thyme-TimeML annotates generic events present in sections documenting doctors' discussion of risks, plans and alternative strategies. They do so because adding these events to a patient's clinical timeline could be important from a legal perspective, as they help to establish informed consent and knowledge of risk. We do not annotate pure generics, because we do not perceive any domain-agnostic utility in annotating them. Note that we annotate verbs of discussion and comprehension which are not generics, so we do not completely ignore events associated with patient consent. For example, in the sentence ``She repeated the potential side effects back to me,'' \emph{repeated} is annotated, but \emph{effects} is not. Thyme-TimeML would have annotated both. \\
\noindent
\textbf{2. Entities as events: }Thyme-TimeML treats some entities and non-events as events in clinical language. Two categories see this shift in semantic interpretation: (i) Medications, and (ii) Disorders. Both categories contribute significant information to a patient's timeline, and so they are treated as events. Since we are not specifically focused on timeline construction, we do not treat these as events. To ensure that we do not discard potentially crucial information, we incorporate an additional step in which we annotate entities such as medications, body parts, abnormalities (e.g., rash), etc.

\begin{table}[]
    \centering
    \begin{tabular}{lll}
       \toprule \textbf{Domain} & \textbf{Entity $\kappa$} & \textbf{Event $\kappa$} \\ \midrule
        \textbf{Notes} & 0.9117 & 0.8652 \\
        \textbf{Convos} & 0.8634 & 0.8327 \\ \bottomrule
    \end{tabular}
    \caption{Inter-annotator agreement on entity and event annotation tasks in both domains, measured using chance-corrected Cohen's $\kappa$}
    \label{tab:agreement}
\end{table}

\subsection{Annotation Process}
\begin{figure}
    \centering
    \includegraphics[scale=0.35]{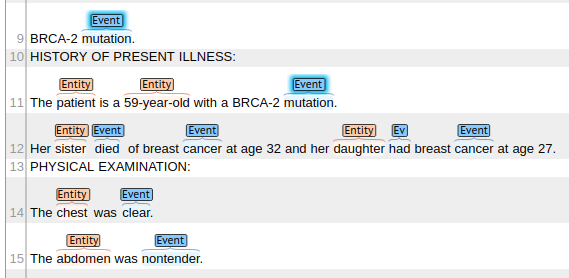}
    \caption{Sample clinical note with entity and event annotation}
    \label{fig:brat}
\end{figure}

After incorporating our modifications, we test our guidelines by having two expert annotators annotate one document from each domain. We observe high inter-annotator agreement (measured by chance-corrected Cohen's $\kappa$) on both entity and event annotation, in both domains. Table~\ref{tab:agreement} presents the agreement scores. To create our final datasets, we sample 45 documents from each domain (15 from each specialty). Each document is annotated by one expert. Annotation is carried out using the BRAT stand-off markup interface \cite{stenetorp-etal-2012-brat}. Figure~\ref{fig:brat} shows a sample clinical note annotated with events and entities. Table~\ref{tab:datastats} gives a brief overview of statistics for our datasets, in comparison with TimeBank (news articles) \cite{pustejovsky2003timebank}.

\begin{table}[]
    \centering
    \begin{tabular}{llll}
    \toprule \textbf{Statistic} & \textbf{News} & \textbf{Notes} & \textbf{Convos}\\ \midrule
    \textbf{\#Files} & 54 & 45 & 45 \\
    \textbf{\#Tokens} & 18,263 & 28,935 & 76,711\\
    \textbf{\#Events} & 1986 & 4781 & 7064\\
    \textbf{Event Density} & 10.88\% & 16.52\% & 9.21\%\\
    \textbf{Vocab Size} & 3978 & 4303 & 3505\\
    \textbf{Event Vocab} & 1015 & 1588 & 1472\\ \bottomrule
    \end{tabular}
    \caption{Dataset statistics. Note that the statistics for TimeBank (News) are computed over the test set for fair comparison with our datasets, which are test-only.}
    \label{tab:datastats}
\end{table}

\section{Methods for Marginal Alignment}
To adapt event extraction models with no training data, we tackle the problem of covariate shift, which arises when the marginal distribution (or input distribution) $P(X)$ changes between train and test data. Directly applying a supervised model trained on the training set, to the test set might not perform well due to the gap between training and test distributions. We experiment with several techniques to align the training and test distributions, so that the supervised model transfers better to test data. They can be divided into two types based on the kind of supervision used during alignment: (i) task-guided alignment techniques, and (ii) task-agnostic alignment techniques

\subsection{Task-Guided Alignment Techniques}
These techniques jointly optimize for two tasks: (i) aligning training and test distributions, and (ii) training an event extraction model. Since the alignment process receives supervision from task training, we refer to these techniques as task-guided alignment techniques. Under this category, we experiment with adversarial domain adaptation.\\
\noindent
\textbf{Adversarial Domain Adaptation:} Adversarial domain adaptation was proposed by \newcite{ganin2015unsupervised}, who showed its efficacy on sentiment analysis. Recently, \newcite{naik2020open} showed that adversarial domain adaptation could be used to transfer event extraction models between two domains: news and literature. The adversarial domain adaptation framework for event extraction contains three components: (i) representation learner ($R$) which generates token-level representations for a sequence, (ii) event classifier which identifies events ($E$), and (iii) domain predictor ($D$) which predicts the domain for the sequence. The key idea is to train $R$ to generate representations which \emph{are predictive} for event identification but \emph{not predictive} for domain prediction, making it more domain-invariant. This aligns training and test distributions by finding a shared feature space in which training and test samples are not distinguishable, while making sure that the feature space is useful for event extraction. The technique relies on an alternating optimization procedure. The first step optimizes $D$ on the domain prediction task, while the second step optimizes both $R$ and $E$ on event identification while subtracting domain prediction loss. For complete mathematical details, we refer the interested reader to \newcite{naik2020open}. 


\subsection{Task-Agnostic Alignment Techniques}
These techniques perform training/test distribution alignment and event extraction training sequentially instead of jointly optimizing them. The alignment process does not receive supervision from task training, so these techniques are task-agnostic. We experiment with the following techniques:

\noindent 
\textbf{Domain Adaptive Fine-tuning:} Domain adaptive fine-tuning has been proposed as an effective technique for unsupervised adaption of sequence labeling models to challenging domains such as Early Modern English and social media \cite{han-eisenstein-2019-unsupervised}. This procedure works as follows:
\vspace{-0.5em}
\begin{enumerate}
\setlength\itemsep{-0.4em}
    \item Create a large dataset containing equal proportions of sentences from source and target domains. Fine-tune contextualized embeddings using a masked language modeling objective.
    \item Using fine-tuned embeddings, train an event extraction model on labeled source data.
\end{enumerate}
In addition to this setup, we experiment with a variant of this procedure, which uses a syntactic objective function. This variant fine-tunes embeddings on the POS tagging task in step 1. The motivation behind this variant is two-fold. First, we observe that event annotation is heavily syntax-driven, allowing delexicalized models (i.e., models using POS tags instead of words) to achieve high performance (\S\ref{ssec:results}). This indicates that infusing more syntactic awareness into embeddings might help performance on the task. Second, syntax might offer an additional basis for generalization, since sentences that look very different lexically, might follow similar syntactic structures. Intuitively, this variant is similar to syntactic relexicalization which has shown success in cross-lingual dependency parsing \cite{duong-etal-2015-cross}. 

\noindent
\textbf{Likelihood-based Instance Weighting:} We develop a novel instance weighting procedure which uses likelihood scores computed by a language model. Instance selection and instance weighting strategies have frequently been used to perform domain adaptation by correcting for distributional differences \cite{jiang-zhai-2007-instance,foster-etal-2010-discriminative,axelrod-etal-2011-domain,wang-etal-2017-instance}. The main premise is that some samples from out-of-domain data and in-domain data often share some characteristics. Training only on these samples (pruning), or biasing training to focus more on these samples (weighting) can produce models that perform better on out-of-domain data. Motivated by this, our instance weighting procedure works as follows.

Let $S_t = w_1 w_2 ... w_n$ be a sentence from the in-domain training set. Let $O$ be a language model trained on raw text from the target domain. We first compute the likelihood of sentence $S_t$ under $O$ as $\mathbb{L}_t = P_O(w_1) \Pi_{i=2}^n P_O(w_i | w_1...w_{i-1})$, where $P_O$ indicates probability under model $O$. Then we compute a weight for $S_t$ as follows:
\begin{align}
    \alpha_{S_t} = \frac{\mathbb{L}_t}{\sum_{i=1}^{|N|} \mathbb{L}_i} * |N|
\end{align}
where $|N|$ is the size of in-domain training set. This metric gives a higher weight to in-domain sentences that are \emph{more likely} under the target domain language model, up-weighting instances that share more characteristics with target domain sentences. The alpha values are used to weight the loss function, thus biasing the training procedure.

\begin{table*}[h]
    \centering
    \begin{tabular}{lcccccc}
    \toprule \textbf{Model} & \multicolumn{3}{c}{\textbf{In-Domain}} & \multicolumn{3}{c}{\textbf{Out-of-Domain}}  \\ \cmidrule{2-7}
     & \textbf{P} & \textbf{R} & \textbf{F1} & \textbf{P} & \textbf{R} & \textbf{F1} \\ \midrule
     \textbf{VERB}& 58.8 & 66.5 & 62.5 & 49.4 & 41.4 & 45.0\\
    \textbf{DELEX} & 75.0 & 66.3 & 70.4 & 74.4 & 42.2 & 53.8\\
    \textbf{BERT} & 80.6 & 86.0 & 83.2 & 85.7 & 55.9 & 67.6\\
    \textbf{CBERT} & 79.2 & 83.3 & 81.2 & 85.8 & 52.9 & 65.4 \\\midrule
    \textbf{BERT-ADA} & 81.2 & 86.3 & 83.7 & 83.2 & \textbf{60.4} & \textbf{70.0}\\
    \textbf{BERT-LIW} & 81.9 & 86.6 & 84.1 & \textbf{86.7} & 56.0 & 68.1\\
    \textbf{BERT-DAFT} & 79.1 & 85.9 & 82.3 & 83.9 & 58.6 & 69.0\\
    \textbf{BERT-DAFT-SYN} & 76.9 & 80.7 & 78.7 & 70.7 & 56.8 & 63.0\\ \bottomrule
    \end{tabular}
    \caption{Model performance on domain transfer experiments from news to clinical notes.}
    \label{tab:notes}
\end{table*}

\begin{table*}[h]
    \centering
    \begin{tabular}{lcccccc}
    \toprule \textbf{Model} & \multicolumn{3}{c}{\textbf{In-Domain}} & \multicolumn{3}{c}{\textbf{Out-of-Domain}}  \\ \cmidrule{2-7}
     & \textbf{P} & \textbf{R} & \textbf{F1} & \textbf{P} & \textbf{R} & \textbf{F1} \\ \midrule
     \textbf{VERB}& 58.8 & 66.5 & 62.5 & 44.6 & 68.1 & 53.9\\
    \textbf{DELEX} & 75.0 & 66.3 & 70.4 & 56.9 & 64.5 & 60.4\\
    \textbf{BERT} & 80.6 & 86.0 & 83.2 & 75.0 & 63.6 & 68.9\\
    \textbf{CBERT} & 79.2 & 83.3 & 81.2 & 66.5 & 65.1 & 65.8 \\\midrule
    \textbf{BERT-ADA} & 81.1 & 85.9 & 83.4 & \textbf{74.5} & 62.2 & 67.8\\
    \textbf{BERT-LIW} & 80.0 & 87.0 & 83.4 & 72.8 & 67.3 & 70.0\\
    \textbf{BERT-DAFT} & 78.5 & 84.8 & 81.5 & 72.7 & \textbf{73.1 }& \textbf{72.9}\\
    \textbf{BERT-DAFT-SYN} & 80.0 & 78.7 & 79.3 & 67.6 & 60.7 & 63.9\\ \bottomrule
    \end{tabular}
    \caption{Model performance on domain transfer experiments from news to doctor-patient conversations.}
    \label{tab:convos}
\end{table*}

\section{Experimental Setup}
\subsection{Model Details}
The goal of our evaluation is to identify which alignment techniques work best when transferring event extraction models to medical data. We choose a strong BERT-based baseline model with no transfer, and evaluate the performance of each alignment technique when applied to this baseline.\\
\noindent
\textbf{VERB:} A simple unsupervised baseline labeling all verbs as events.\\
\textbf{DELEX:} A fully-delexicalized baseline using POS tag embeddings as features, followed by an MLP.\\ 
\textbf{BERT:} A single-layer BiLSTM over contextual embeddings extracted using BERT \cite{devlin-etal-2019-bert}, followed by an MLP, similar to the best-performing model on LitBank \cite{sims-etal-2019-literary}.\\
\textbf{CBERT:} Similar to BERT, but embeddings are extracted from Clinical-BERT \cite{alsentzer-etal-2019-publicly}\\
\textbf{BERT-ADA:} BERT baseline trained using adversarial domain adaptation. \\
\textbf{BERT-LIW:} BERT baseline trained on data weighted by LM likelihood. We train autoregressive language models using 3 million tokens for each target domain.\\
\textbf{BERT-DAFT:} BERT baseline with domain adaptive fine-tuning. We use the same target domain text used to train LMs for BERT-LIW. For news, we extract 3 million tokens from the CNN/ DailyMail dataset \cite{hermann2015teaching}.\\
\textbf{BERT-DAFT-SYN:} BERT baseline with syntactic fine-tuning on the same source+target text as BERT-DAFT, POS tagged using Stanford CoreNLP \cite{manning-etal-2014-stanford}.
\subsection{Results}
\label{ssec:results}
Tables~\ref{tab:notes} and~\ref{tab:convos} show the performance of all models when transferring from news data to clinical notes and doctor-patient conversations respectively. From the tables, we see that DELEX is surprisingly strong out-of-domain. BERT with no transfer performs well out-of-domain, improving by 8.25 F1 points on average over DELEX. C-BERT also performs well out-of-domain, but does worse than BERT. We attribute this to the fact that fine-tuning only on clinical notes does not improve alignment with the source domain, providing no basis for models trained on news to adapt better. BERT-ADA shows mixed results, improving over BERT by 2.4 F1 on notes, but dropping by 1.1 F1 on conversations. BERT-LIW and BERT-DAFT improve upon BERT in both domains. BERT-DAFT shows minor performance drops in-domain, due to some degree of catastrophic forgetting. BERT-DAFT-SYN shows performance drops, both in-domain and out-of-domain, in both settings. Unlike syntactic relexicalization work which used non-contextualized embeddings, we use contextualized embeddings, which possess a larger degree of syntactic information, probably reducing the need for syntax-driven training. Another source of errors is POS tagging, since off-the-shelf taggers trained on news will be less accurate on our data. Across domains, the skew between precision and recall is higher on notes, which might stem from the specialized vocabulary used in them dragging down recall.   
\section{Analysis and Discussion}
\label{sec:disc}
Tables~\ref{tab:notes} and~\ref{tab:convos} provide an indication of model ability to handle covariate shift. However, covariate shift occurs at multiple layers in language (e.g., lexical level, syntactic level, etc.), leading to different dimensions of variation between domains (e.g., topical variation, genre variation, etc.). Looking at overall model performance does not offer insight into whether there are specific shifts that some models are better at addressing. We dig deeper into this question, focusing on two levels of shift: (i) lexical shift, and (ii) semantic (event type) shift.\\
\noindent
\textbf{Variation under lexical shift:} We separate model performance on in-vocabulary (IV) and out-of-vocabulary (OOV) tokens. Note that the proportion of events that are OOV is higher in clinical notes (52\%) than conversations (20.6\%). Tables~\ref{tab:oovnotes} and~\ref{tab:oovconvos} present model performance on these token categories. Surprisingly, despite the use of specialized language, OOV performance on clinical notes is higher than conversations for all models except BERT-DAFT. Taking a closer look at the OOV event instances from clinical notes that models identify correctly, we see that a large proportion (54.8\%) contain one of three morphological patterns: (i) past tense verbs ending in ``-ed'', (ii) gerunds ending in ``-ing'', or (iii) nouns ending in ``-tion'' or ``-sion''. These patterns are also common among events in the news domain. For example, past tense verbs often refer to events that have already occurred and gerunds and nouns ending in ``-tion'' refer to processes. We hypothesize that BERT-based models might be exploiting these morphological regularities to correctly label unseen medical terms (e.g., irrigated, excision, dissected, wheezing, etc.). These patterns are more prevalent in notes (35.6\%) than conversations (23.5\%), explaining the surprising performance difference.

\begin{table}[]
    \centering
    \begin{tabular}{lcc}
     \toprule \textbf{Model} & \textbf{IV F1} & \textbf{OOV F1}\\ \midrule
    \textbf{BERT} & 73.5 & 61.2 \\
    \textbf{BERT-ADA} & 75.2 & 65.0 \\
    \textbf{BERT-LIW} & 73.6 & 62.6 \\
    \textbf{BERT-DAFT} & 75.7 & 62.0 \\
    \textbf{BERT-DAFT-SYN} & 67.7 & 58.4 \\ \bottomrule
    \end{tabular}
    \caption{Model performance on in-vocabulary (IV) and out-of-vocabulary (OOV) terms from clinical notes.}
    \label{tab:oovnotes}
\end{table}
\begin{table}[]
    \centering
    \begin{tabular}{lcc}
     \toprule \textbf{Model} & \textbf{IV F1} & \textbf{OOV F1}\\ \midrule
    \textbf{BERT} & 71.3 & 57.9 \\
    \textbf{BERT-ADA} & 70.2 & 57.6 \\
    \textbf{BERT-LIW} & 72.0 & 61.4 \\
    \textbf{BERT-DAFT} & 74.9 & 63.6 \\
    \textbf{BERT-DAFT-SYN} & 65.5 & 55.5 \\ \bottomrule
    \end{tabular}
    \caption{Model performance on in-vocabulary (IV) and out-of-vocabulary (OOV) terms from doctor-patient conversations.}
    \label{tab:oovconvos}
\end{table}
\noindent
\textbf{Variation under semantic shift:} To determine whether model performance on OOV tokens depends on event type, we randomly sample $\sim$500 OOV tokens from each domain and label them for event type. We use the same typology as TimeML (State, I-State, Occurrence, Aspectual, None), with labels for the event types we introduce (ActivityPattern, LongTermState). We run an ANOVA model with each token per model as an instance (total 5080 instances), noting Event Type, Target (notes/convos), Model (BERT/ADA/LIW/DAFT/DAFT-SYN) and Correctness (1 vs 0). Correctness is the dependent variable, while others are independent variables. We include all pairwise interaction terms and the three way interaction between Event Type, Target and Model. We see a positive main effect of Event Type on Correctness ($p<0.0001$), indicating that some event types are more difficult. There are two significant two-way interactions, one between Target and Event type ($p<0.0001$), indicating that difficulty of event types differs across sources, and between Model and Event type ($p<0.0001$), indicating that which model is better depends on event type. Three way interaction between Model, Event type, and Target is also significant ($p<0.0001$), indicating that performance differences between models per event type differs between sources. 

We interpret differences in performance per event type separately for each source using a student-t post-hoc analysis to determine which pairwise contrasts are statistically significant. This reveals that in clinical notes, LIW outperforms all models on I-State events (i.e., hypothetical, future or negated states) and LongTermState events, a category never seen in the training data. These improvements might stem from the training algorithm used by LIW. LIW up-weights instances in news that resemble medical data, which contains a high proportion of these event categories. Therefore, despite being infrequent in news, they get up-weighted, helping LIW identify them better.

\section{Conclusion}
In this work, we focused on unsupervised adaptation of event extractors to new domains by aligning the marginal distributions of source and target domains. We created two event extraction test sets using English texts from two medical domains: (i) clinical notes, and (ii) doctor-patient conversations, and tested the efficacy of three alignment techniques: (i) adversarial domain adaptation (ADA), (ii) domain adaptive fine-tuning (DAFT), and (iii) a novel instance weighting technique based on language model likelihood scores (LIW). None of these models consistently outperformed the others, but a deeper analysis of model performance under different types of shifts (e.g., lexical shift, semantic shift) uncovered interesting variations among models. Our best-performing models attained F1 scores of 70.0 and 72.9 on notes and conversations respectively, using \emph{no} labeled target data. We believe these models define a good starting point and can be further improved using few-shot learning.

\bibliographystyle{acl_natbib}
\bibliography{emnlp2020}

\begin{thebibliography}{55}
\expandafter\ifx\csname natexlab\endcsname\relax\def\natexlab#1{#1}\fi

\bibitem[{Alsentzer et~al.(2019)Alsentzer, Murphy, Boag, Weng, Jindi, Naumann,
  and McDermott}]{alsentzer-etal-2019-publicly}
Emily Alsentzer, John Murphy, William Boag, Wei-Hung Weng, Di~Jindi, Tristan
  Naumann, and Matthew McDermott. 2019.
\newblock \href {https://doi.org/10.18653/v1/W19-1909} {Publicly available
  clinical {BERT} embeddings}.
\newblock In \emph{Proceedings of the 2nd Clinical Natural Language Processing
  Workshop}, pages 72--78, Minneapolis, Minnesota, USA. Association for
  Computational Linguistics.

\bibitem[{Araki and Mitamura(2018)}]{araki-mitamura-2018-open}
Jun Araki and Teruko Mitamura. 2018.
\newblock \href {https://www.aclweb.org/anthology/C18-1075} {Open-domain event
  detection using distant supervision}.
\newblock In \emph{Proceedings of the 27th International Conference on
  Computational Linguistics}, pages 878--891, Santa Fe, New Mexico, USA.
  Association for Computational Linguistics.

\bibitem[{Axelrod et~al.(2011)Axelrod, He, and Gao}]{axelrod-etal-2011-domain}
Amittai Axelrod, Xiaodong He, and Jianfeng Gao. 2011.
\newblock \href {https://www.aclweb.org/anthology/D11-1033} {Domain adaptation
  via pseudo in-domain data selection}.
\newblock In \emph{Proceedings of the 2011 Conference on Empirical Methods in
  Natural Language Processing}, pages 355--362, Edinburgh, Scotland, UK.
  Association for Computational Linguistics.

\bibitem[{Bach(1986)}]{bach1986algebra}
Emmon Bach. 1986.
\newblock The algebra of events.
\newblock \emph{Linguistics and philosophy}, pages 5--16.

\bibitem[{Berant et~al.(2014)Berant, Srikumar, Chen, Vander~Linden, Harding,
  Huang, Clark, and Manning}]{berant-etal-2014-modeling}
Jonathan Berant, Vivek Srikumar, Pei-Chun Chen, Abby Vander~Linden, Brittany
  Harding, Brad Huang, Peter Clark, and Christopher~D. Manning. 2014.
\newblock \href {https://doi.org/10.3115/v1/D14-1159} {Modeling biological
  processes for reading comprehension}.
\newblock In \emph{Proceedings of the 2014 Conference on Empirical Methods in
  Natural Language Processing ({EMNLP})}, pages 1499--1510, Doha, Qatar.
  Association for Computational Linguistics.

\bibitem[{Bethard et~al.(2015)Bethard, Derczynski, Savova, Pustejovsky, and
  Verhagen}]{bethard-etal-2015-semeval}
Steven Bethard, Leon Derczynski, Guergana Savova, James Pustejovsky, and Marc
  Verhagen. 2015.
\newblock \href {https://doi.org/10.18653/v1/S15-2136} {{S}em{E}val-2015 task
  6: Clinical {T}emp{E}val}.
\newblock In \emph{Proceedings of the 9th International Workshop on Semantic
  Evaluation ({S}em{E}val 2015)}, pages 806--814, Denver, Colorado. Association
  for Computational Linguistics.

\bibitem[{Bethard et~al.(2016)Bethard, Savova, Chen, Derczynski, Pustejovsky,
  and Verhagen}]{bethard-etal-2016-semeval}
Steven Bethard, Guergana Savova, Wei-Te Chen, Leon Derczynski, James
  Pustejovsky, and Marc Verhagen. 2016.
\newblock \href {https://doi.org/10.18653/v1/S16-1165} {{S}em{E}val-2016 task
  12: Clinical {T}emp{E}val}.
\newblock In \emph{Proceedings of the 10th International Workshop on Semantic
  Evaluation ({S}em{E}val-2016)}, pages 1052--1062, San Diego, California.
  Association for Computational Linguistics.

\bibitem[{Bethard et~al.(2017)Bethard, Savova, Palmer, and
  Pustejovsky}]{bethard-etal-2017-semeval}
Steven Bethard, Guergana Savova, Martha Palmer, and James Pustejovsky. 2017.
\newblock \href {https://doi.org/10.18653/v1/S17-2093} {{S}em{E}val-2017 task
  12: Clinical {T}emp{E}val}.
\newblock In \emph{Proceedings of the 11th International Workshop on Semantic
  Evaluation ({S}em{E}val-2017)}, pages 565--572, Vancouver, Canada.
  Association for Computational Linguistics.

\bibitem[{Blitzer et~al.(2007)Blitzer, Dredze, and
  Pereira}]{blitzer-etal-2007-biographies}
John Blitzer, Mark Dredze, and Fernando Pereira. 2007.
\newblock \href {https://www.aclweb.org/anthology/P07-1056} {Biographies,
  {B}ollywood, boom-boxes and blenders: Domain adaptation for sentiment
  classification}.
\newblock In \emph{Proceedings of the 45th Annual Meeting of the Association of
  Computational Linguistics}, pages 440--447, Prague, Czech Republic.
  Association for Computational Linguistics.

\bibitem[{Blitzer et~al.(2006)Blitzer, McDonald, and
  Pereira}]{blitzer-etal-2006-domain}
John Blitzer, Ryan McDonald, and Fernando Pereira. 2006.
\newblock \href {https://www.aclweb.org/anthology/W06-1615} {Domain adaptation
  with structural correspondence learning}.
\newblock In \emph{Proceedings of the 2006 Conference on Empirical Methods in
  Natural Language Processing}, pages 120--128, Sydney, Australia. Association
  for Computational Linguistics.

\bibitem[{Chambers et~al.(2014)Chambers, Cassidy, McDowell, and
  Bethard}]{chambers-etal-2014-dense}
Nathanael Chambers, Taylor Cassidy, Bill McDowell, and Steven Bethard. 2014.
\newblock \href {https://doi.org/10.1162/tacl_a_00182} {Dense event ordering
  with a multi-pass architecture}.
\newblock \emph{Transactions of the Association for Computational Linguistics},
  2:273--284.

\bibitem[{Chen et~al.(2014)Chen, Weinberger, Sha, and
  Bengio}]{chen2014marginalized}
Minmin Chen, Kilian~Q. Weinberger, Fei Sha, and Yoshua Bengio. 2014.
\newblock \href {http://proceedings.mlr.press/v32/cheng14.html} {Marginalized
  denoising auto-encoders for nonlinear representations}.
\newblock In \emph{Proceedings of the 31th International Conference on Machine
  Learning, {ICML} 2014, Beijing, China, 21-26 June 2014}, volume~32 of
  \emph{{JMLR} Workshop and Conference Proceedings}, pages 1476--1484.
  JMLR.org.

\bibitem[{Cocos et~al.(2017)Cocos, Fiks, and Masino}]{cocos2017deep}
Anne Cocos, Alexander~G. Fiks, and Aaron~J. Masino. 2017.
\newblock \href {https://doi.org/10.1093/jamia/ocw180} {Deep learning for
  pharmacovigilance: recurrent neural network architectures for labeling
  adverse drug reactions in twitter posts}.
\newblock \emph{J. Am. Medical Informatics Assoc.}, 24(4):813--821.

\bibitem[{Cybulska and Vossen(2014)}]{cybulska-vossen-2014-using}
Agata Cybulska and Piek Vossen. 2014.
\newblock \href
  {http://www.lrec-conf.org/proceedings/lrec2014/pdf/840_Paper.pdf} {Using a
  sledgehammer to crack a nut? lexical diversity and event coreference
  resolution}.
\newblock In \emph{Proceedings of the Ninth International Conference on
  Language Resources and Evaluation ({LREC}'14)}, pages 4545--4552, Reykjavik,
  Iceland. European Language Resources Association (ELRA).

\bibitem[{Devlin et~al.(2019)Devlin, Chang, Lee, and
  Toutanova}]{devlin-etal-2019-bert}
Jacob Devlin, Ming-Wei Chang, Kenton Lee, and Kristina Toutanova. 2019.
\newblock \href {https://doi.org/10.18653/v1/N19-1423} {{BERT}: Pre-training of
  deep bidirectional transformers for language understanding}.
\newblock In \emph{Proceedings of the 2019 Conference of the North {A}merican
  Chapter of the Association for Computational Linguistics: Human Language
  Technologies, Volume 1 (Long and Short Papers)}, pages 4171--4186,
  Minneapolis, Minnesota. Association for Computational Linguistics.

\bibitem[{Doddington et~al.(2004)Doddington, Mitchell, Przybocki, Ramshaw,
  Strassel, and Weischedel}]{doddington2004automatic}
George Doddington, Alexis Mitchell, Mark Przybocki, Lance Ramshaw, Stephanie
  Strassel, and Ralph Weischedel. 2004.
\newblock \href {http://www.lrec-conf.org/proceedings/lrec2004/pdf/5.pdf} {The
  automatic content extraction ({ACE}) program {--} tasks, data, and
  evaluation}.
\newblock In \emph{Proceedings of the Fourth International Conference on
  Language Resources and Evaluation ({LREC}{'}04)}, Lisbon, Portugal. European
  Language Resources Association (ELRA).

\bibitem[{Duong et~al.(2015)Duong, Cohn, Bird, and
  Cook}]{duong-etal-2015-cross}
Long Duong, Trevor Cohn, Steven Bird, and Paul Cook. 2015.
\newblock \href {https://doi.org/10.18653/v1/K15-1012} {Cross-lingual transfer
  for unsupervised dependency parsing without parallel data}.
\newblock In \emph{Proceedings of the Nineteenth Conference on Computational
  Natural Language Learning}, pages 113--122, Beijing, China. Association for
  Computational Linguistics.

\bibitem[{Foster et~al.(2010)Foster, Goutte, and
  Kuhn}]{foster-etal-2010-discriminative}
George Foster, Cyril Goutte, and Roland Kuhn. 2010.
\newblock \href {https://www.aclweb.org/anthology/D10-1044} {Discriminative
  instance weighting for domain adaptation in statistical machine translation}.
\newblock In \emph{Proceedings of the 2010 Conference on Empirical Methods in
  Natural Language Processing}, pages 451--459, Cambridge, MA. Association for
  Computational Linguistics.

\bibitem[{Ganin and Lempitsky(2015)}]{ganin2015unsupervised}
Yaroslav Ganin and Victor~S. Lempitsky. 2015.
\newblock \href {http://proceedings.mlr.press/v37/ganin15.html} {Unsupervised
  domain adaptation by backpropagation}.
\newblock In \emph{Proceedings of the 32nd International Conference on Machine
  Learning, {ICML} 2015, Lille, France, 6-11 July 2015}, volume~37 of
  \emph{{JMLR} Workshop and Conference Proceedings}, pages 1180--1189.
  JMLR.org.

\bibitem[{Ganin et~al.(2016)Ganin, Ustinova, Ajakan, Germain, Larochelle,
  Laviolette, Marchand, and Lempitsky}]{ganin2016domain}
Yaroslav Ganin, Evgeniya Ustinova, Hana Ajakan, Pascal Germain, Hugo
  Larochelle, Fran{\c{c}}ois Laviolette, Mario Marchand, and Victor~S.
  Lempitsky. 2016.
\newblock \href {http://jmlr.org/papers/v17/15-239.html} {Domain-adversarial
  training of neural networks}.
\newblock \emph{J. Mach. Learn. Res.}, 17:59:1--59:35.

\bibitem[{Glorot et~al.(2011)Glorot, Bordes, and Bengio}]{glorot2011domain}
Xavier Glorot, Antoine Bordes, and Yoshua Bengio. 2011.
\newblock \href {https://icml.cc/2011/papers/342\_icmlpaper.pdf} {Domain
  adaptation for large-scale sentiment classification: {A} deep learning
  approach}.
\newblock In \emph{Proceedings of the 28th International Conference on Machine
  Learning, {ICML} 2011, Bellevue, Washington, USA, June 28 - July 2, 2011},
  pages 513--520. Omnipress.

\bibitem[{Grishman and Sundheim(1996)}]{grishman-sundheim-1996-message}
Ralph Grishman and Beth Sundheim. 1996.
\newblock \href {https://www.aclweb.org/anthology/C96-1079} {Message
  understanding conference- 6: A brief history}.
\newblock In \emph{{COLING} 1996 Volume 1: The 16th International Conference on
  Computational Linguistics}.

\bibitem[{Gui et~al.(2017)Gui, Zhang, Huang, Peng, and
  Huang}]{gui-etal-2017-part}
Tao Gui, Qi~Zhang, Haoran Huang, Minlong Peng, and Xuanjing Huang. 2017.
\newblock \href {https://doi.org/10.18653/v1/D17-1256} {Part-of-speech tagging
  for twitter with adversarial neural networks}.
\newblock In \emph{Proceedings of the 2017 Conference on Empirical Methods in
  Natural Language Processing}, pages 2411--2420, Copenhagen, Denmark.
  Association for Computational Linguistics.

\bibitem[{Gururangan et~al.(2020)Gururangan, Marasović, Swayamdipta, Lo,
  Beltagy, Downey, and Smith}]{gururangan2020dont}
Suchin Gururangan, Ana Marasović, Swabha Swayamdipta, Kyle Lo, Iz~Beltagy,
  Doug Downey, and Noah~A. Smith. 2020.
\newblock \href {http://arxiv.org/abs/2004.10964} {Don't stop pretraining:
  Adapt language models to domains and tasks}.

\bibitem[{Han and Eisenstein(2019)}]{han-eisenstein-2019-unsupervised}
Xiaochuang Han and Jacob Eisenstein. 2019.
\newblock \href {https://doi.org/10.18653/v1/D19-1433} {Unsupervised domain
  adaptation of contextualized embeddings for sequence labeling}.
\newblock In \emph{Proceedings of the 2019 Conference on Empirical Methods in
  Natural Language Processing and the 9th International Joint Conference on
  Natural Language Processing (EMNLP-IJCNLP)}, pages 4238--4248, Hong Kong,
  China. Association for Computational Linguistics.

\bibitem[{Henry et~al.(2020)Henry, Buchan, Filannino, Stubbs, and
  Uzuner}]{henry2018n2c2}
Sam Henry, Kevin Buchan, Michele Filannino, Amber Stubbs, and {\"{O}}zlem
  Uzuner. 2020.
\newblock \href {https://doi.org/10.1093/jamia/ocz166} {2018 n2c2 shared task
  on adverse drug events and medication extraction in electronic health
  records}.
\newblock \emph{J. Am. Medical Informatics Assoc.}, 27(1):3--12.

\bibitem[{Hermann et~al.(2015)Hermann, Kocisk{\'{y}}, Grefenstette, Espeholt,
  Kay, Suleyman, and Blunsom}]{hermann2015teaching}
Karl~Moritz Hermann, Tom{\'{a}}s Kocisk{\'{y}}, Edward Grefenstette, Lasse
  Espeholt, Will Kay, Mustafa Suleyman, and Phil Blunsom. 2015.
\newblock \href
  {http://papers.nips.cc/paper/5945-teaching-machines-to-read-and-comprehend}
  {Teaching machines to read and comprehend}.
\newblock In \emph{Advances in Neural Information Processing Systems 28: Annual
  Conference on Neural Information Processing Systems 2015, December 7-12,
  2015, Montreal, Quebec, Canada}, pages 1693--1701.

\bibitem[{Jain et~al.(2016)Jain, Kasiviswanathan, and
  Huang}]{jain-etal-2016-towards}
Ajit Jain, Girish Kasiviswanathan, and Ruihong Huang. 2016.
\newblock \href {https://www.aclweb.org/anthology/W16-3911} {Towards accurate
  event detection in social media: A weakly supervised approach for learning
  implicit event indicators}.
\newblock In \emph{Proceedings of the 2nd Workshop on Noisy User-generated Text
  ({WNUT})}, pages 70--77, Osaka, Japan. The COLING 2016 Organizing Committee.

\bibitem[{Jiang and Zhai(2007)}]{jiang-zhai-2007-instance}
Jing Jiang and ChengXiang Zhai. 2007.
\newblock \href {https://www.aclweb.org/anthology/P07-1034} {Instance weighting
  for domain adaptation in {NLP}}.
\newblock In \emph{Proceedings of the 45th Annual Meeting of the Association of
  Computational Linguistics}, pages 264--271, Prague, Czech Republic.
  Association for Computational Linguistics.

\bibitem[{Keith et~al.(2017)Keith, Handler, Pinkham, Magliozzi, McDuffie, and
  O{'}Connor}]{keith-etal-2017-identifying}
Katherine Keith, Abram Handler, Michael Pinkham, Cara Magliozzi, Joshua
  McDuffie, and Brendan O{'}Connor. 2017.
\newblock \href {https://doi.org/10.18653/v1/D17-1163} {Identifying civilians
  killed by police with distantly supervised entity-event extraction}.
\newblock In \emph{Proceedings of the 2017 Conference on Empirical Methods in
  Natural Language Processing}, pages 1547--1557, Copenhagen, Denmark.
  Association for Computational Linguistics.

\bibitem[{Kim et~al.(2009)Kim, Ohta, Pyysalo, Kano, and
  Tsujii}]{kim-etal-2009-overview}
Jin-Dong Kim, Tomoko Ohta, Sampo Pyysalo, Yoshinobu Kano, and Jun{'}ichi
  Tsujii. 2009.
\newblock \href {https://www.aclweb.org/anthology/W09-1401} {Overview of
  {B}io{NLP}{'}09 shared task on event extraction}.
\newblock In \emph{Proceedings of the {B}io{NLP} 2009 Workshop Companion Volume
  for Shared Task}, pages 1--9, Boulder, Colorado. Association for
  Computational Linguistics.

\bibitem[{Kim et~al.(2008)Kim, Ohta, and Tsujii}]{kim2008corpus}
Jin{-}Dong Kim, Tomoko Ohta, and Jun'ichi Tsujii. 2008.
\newblock \href {https://doi.org/10.1186/1471-2105-9-10} {Corpus annotation for
  mining biomedical events from literature}.
\newblock \emph{{BMC} Bioinform.}, 9.

\bibitem[{Lee et~al.(2012)Lee, Recasens, Chang, Surdeanu, and
  Jurafsky}]{lee-etal-2012-joint}
Heeyoung Lee, Marta Recasens, Angel Chang, Mihai Surdeanu, and Dan Jurafsky.
  2012.
\newblock \href {https://www.aclweb.org/anthology/D12-1045} {Joint entity and
  event coreference resolution across documents}.
\newblock In \emph{Proceedings of the 2012 Joint Conference on Empirical
  Methods in Natural Language Processing and Computational Natural Language
  Learning}, pages 489--500, Jeju Island, Korea. Association for Computational
  Linguistics.

\bibitem[{Li et~al.(2014)Li, Ritter, Cardie, and Hovy}]{li-etal-2014-major}
Jiwei Li, Alan Ritter, Claire Cardie, and Eduard Hovy. 2014.
\newblock \href {https://doi.org/10.3115/v1/D14-1214} {Major life event
  extraction from twitter based on congratulations/condolences speech acts}.
\newblock In \emph{Proceedings of the 2014 Conference on Empirical Methods in
  Natural Language Processing ({EMNLP})}, pages 1997--2007, Doha, Qatar.
  Association for Computational Linguistics.

\bibitem[{Manning et~al.(2014)Manning, Surdeanu, Bauer, Finkel, Bethard, and
  McClosky}]{manning-etal-2014-stanford}
Christopher Manning, Mihai Surdeanu, John Bauer, Jenny Finkel, Steven Bethard,
  and David McClosky. 2014.
\newblock \href {https://doi.org/10.3115/v1/P14-5010} {The {S}tanford
  {C}ore{NLP} natural language processing toolkit}.
\newblock In \emph{Proceedings of 52nd Annual Meeting of the Association for
  Computational Linguistics: System Demonstrations}, pages 55--60, Baltimore,
  Maryland. Association for Computational Linguistics.

\bibitem[{McCarthy(2002)}]{mccarthy2002actions}
John McCarthy. 2002.
\newblock Actions and other events in situation calculus.
\newblock In \emph{Proceedings of the Eights International Conference on
  Principles and Knowledge Representation and Reasoning (KR-02), Toulouse,
  France, April 22-25, 2002}, pages 615--628. Morgan Kaufmann.

\bibitem[{Mitamura et~al.(2016)Mitamura, Liu, and Hovy}]{mitamura2016overview}
Teruko Mitamura, Zhengzhong Liu, and Eduard~H. Hovy. 2016.
\newblock \href
  {https://tac.nist.gov/publications/2016/additional.papers/TAC2016.KBP\_Event\_Nugget\_overview.proceedings.pdf}
  {Overview of {TAC-KBP} 2016 event nugget track}.
\newblock In \emph{Proceedings of the 2016 Text Analysis Conference, {TAC}
  2016, Gaithersburg, Maryland, USA, November 14-15, 2016}. {NIST}.

\bibitem[{Mitamura et~al.(2015)Mitamura, Yamakawa, Holm, Song, Bies, Kulick,
  and Strassel}]{mitamura-etal-2015-event}
Teruko Mitamura, Yukari Yamakawa, Susan Holm, Zhiyi Song, Ann Bies, Seth
  Kulick, and Stephanie Strassel. 2015.
\newblock \href {https://doi.org/10.3115/v1/W15-0809} {Event nugget annotation:
  Processes and issues}.
\newblock In \emph{Proceedings of the The 3rd Workshop on {EVENTS}: Definition,
  Detection, Coreference, and Representation}, pages 66--76, Denver, Colorado.
  Association for Computational Linguistics.

\bibitem[{Naik et~al.(2017)Naik, Bogart, and Rose}]{naik-etal-2017-extracting}
Aakanksha Naik, Chris Bogart, and Carolyn Rose. 2017.
\newblock \href {https://doi.org/10.18653/v1/W17-2346} {Extracting personal
  medical events for user timeline construction using minimal supervision}.
\newblock In \emph{{B}io{NLP} 2017}, pages 356--364, Vancouver, Canada,.
  Association for Computational Linguistics.

\bibitem[{Naik and Rosé(2020)}]{naik2020open}
Aakanksha Naik and Carolyn Rosé. 2020.
\newblock \href {http://arxiv.org/abs/2005.11355} {Towards open domain event
  trigger identification using adversarial domain adaptation}.

\bibitem[{Nikfarjam et~al.(2015)Nikfarjam, Sarker, O'Connor, Ginn, and
  Gonzalez{-}Hernandez}]{nikfarjam2015pharmacovigilance}
Azadeh Nikfarjam, Abeed Sarker, Karen O'Connor, Rachel~E. Ginn, and Graciela
  Gonzalez{-}Hernandez. 2015.
\newblock \href {https://doi.org/10.1093/jamia/ocu041} {Pharmacovigilance from
  social media: mining adverse drug reaction mentions using sequence labeling
  with word embedding cluster features}.
\newblock \emph{J. Am. Medical Informatics Assoc.}, 22(3):671--681.

\bibitem[{O{'}Gorman et~al.(2016)O{'}Gorman, Wright-Bettner, and
  Palmer}]{ogorman-etal-2016-richer}
Tim O{'}Gorman, Kristin Wright-Bettner, and Martha Palmer. 2016.
\newblock \href {https://doi.org/10.18653/v1/W16-5706} {Richer event
  description: Integrating event coreference with temporal, causal and bridging
  annotation}.
\newblock In \emph{Proceedings of the 2nd Workshop on Computing News Storylines
  ({CNS} 2016)}, pages 47--56, Austin, Texas. Association for Computational
  Linguistics.

\bibitem[{Onyshkevych et~al.(1993)Onyshkevych, Okurowski, and
  Carlson}]{onyshkevych-etal-1993-tasks-domains}
Boyan Onyshkevych, Mary~Ellen Okurowski, and Lynn Carlson. 1993.
\newblock \href {https://doi.org/10.3115/1119149.1119165} {Tasks, domains, and
  languages for information extraction}.
\newblock In \emph{TIPSTER TEXT PROGRAM: PHASE {I}: Proceedings of a Workshop
  held at Fredricksburg, Virginia, September 19-23, 1993}, pages 123--133,
  Fredericksburg, Virginia, USA. Association for Computational Linguistics.

\bibitem[{Pustejovsky et~al.(2003{\natexlab{a}})Pustejovsky, Casta{\~{n}}o,
  Ingria, Saur{\'{\i}}, Gaizauskas, Setzer, Katz, and
  Radev}]{pustejovsky2003timeml}
James Pustejovsky, Jos{\'{e}}~M. Casta{\~{n}}o, Robert Ingria, Roser
  Saur{\'{\i}}, Robert~J. Gaizauskas, Andrea Setzer, Graham Katz, and
  Dragomir~R. Radev. 2003{\natexlab{a}}.
\newblock Timeml: Robust specification of event and temporal expressions in
  text.
\newblock In \emph{New Directions in Question Answering, Papers from 2003
  {AAAI} Spring Symposium, Stanford University, Stanford, CA, {USA}}, pages
  28--34. {AAAI} Press.

\bibitem[{Pustejovsky et~al.(2003{\natexlab{b}})Pustejovsky, Hanks, Sauri, See,
  Gaizauskas, Setzer, Radev, Sundheim, Day, Ferro
  et~al.}]{pustejovsky2003timebank}
James Pustejovsky, Patrick Hanks, Roser Sauri, Andrew See, Robert Gaizauskas,
  Andrea Setzer, Dragomir Radev, Beth Sundheim, David Day, Lisa Ferro, et~al.
  2003{\natexlab{b}}.
\newblock The timebank corpus.
\newblock In \emph{Corpus linguistics}, volume 2003, page~40. Lancaster, UK.

\bibitem[{Ritter et~al.(2012)Ritter, Mausam, Etzioni, and
  Clark}]{ritter2012open}
Alan Ritter, Mausam, Oren Etzioni, and Sam Clark. 2012.
\newblock \href {https://doi.org/10.1145/2339530.2339704} {Open domain event
  extraction from twitter}.
\newblock In \emph{The 18th {ACM} {SIGKDD} International Conference on
  Knowledge Discovery and Data Mining, {KDD} '12, Beijing, China, August 12-16,
  2012}, pages 1104--1112. {ACM}.

\bibitem[{Sarker and Gonzalez(2015)}]{sarker2015portable}
Abeed Sarker and Graciela Gonzalez. 2015.
\newblock \href {https://doi.org/10.1016/j.jbi.2014.11.002} {Portable automatic
  text classification for adverse drug reaction detection via multi-corpus
  training}.
\newblock \emph{J. Biomed. Informatics}, 53:196--207.

\bibitem[{Saur{\'\i} et~al.(2005)Saur{\'\i}, Knippen, Verhagen, and
  Pustejovsky}]{sauri-etal-2005-evita}
Roser Saur{\'\i}, Robert Knippen, Marc Verhagen, and James Pustejovsky. 2005.
\newblock \href {https://www.aclweb.org/anthology/H05-1088} {{E}vita: A robust
  event recognizer for {QA} systems}.
\newblock In \emph{Proceedings of Human Language Technology Conference and
  Conference on Empirical Methods in Natural Language Processing}, pages
  700--707, Vancouver, British Columbia, Canada. Association for Computational
  Linguistics.

\bibitem[{Sims et~al.(2019)Sims, Park, and Bamman}]{sims-etal-2019-literary}
Matthew Sims, Jong~Ho Park, and David Bamman. 2019.
\newblock \href {https://doi.org/10.18653/v1/P19-1353} {Literary event
  detection}.
\newblock In \emph{Proceedings of the 57th Annual Meeting of the Association
  for Computational Linguistics}, pages 3623--3634, Florence, Italy.
  Association for Computational Linguistics.

\bibitem[{Stenetorp et~al.(2012)Stenetorp, Pyysalo, Topi{\'c}, Ohta, Ananiadou,
  and Tsujii}]{stenetorp-etal-2012-brat}
Pontus Stenetorp, Sampo Pyysalo, Goran Topi{\'c}, Tomoko Ohta, Sophia
  Ananiadou, and Jun{'}ichi Tsujii. 2012.
\newblock \href {https://www.aclweb.org/anthology/E12-2021} {brat: a web-based
  tool for {NLP}-assisted text annotation}.
\newblock In \emph{Proceedings of the Demonstrations at the 13th Conference of
  the {E}uropean Chapter of the Association for Computational Linguistics},
  pages 102--107, Avignon, France. Association for Computational Linguistics.

\bibitem[{Styler~IV et~al.(2014)Styler~IV, Bethard, Finan, Palmer, Pradhan,
  de~Groen, Erickson, Miller, Lin, Savova, and
  Pustejovsky}]{styler-iv-etal-2014-temporal}
William~F. Styler~IV, Steven Bethard, Sean Finan, Martha Palmer, Sameer
  Pradhan, Piet~C de~Groen, Brad Erickson, Timothy Miller, Chen Lin, Guergana
  Savova, and James Pustejovsky. 2014.
\newblock \href {https://doi.org/10.1162/tacl_a_00172} {Temporal annotation in
  the clinical domain}.
\newblock \emph{Transactions of the Association for Computational Linguistics},
  2:143--154.

\bibitem[{Wang et~al.(2017)Wang, Utiyama, Liu, Chen, and
  Sumita}]{wang-etal-2017-instance}
Rui Wang, Masao Utiyama, Lemao Liu, Kehai Chen, and Eiichiro Sumita. 2017.
\newblock \href {https://doi.org/10.18653/v1/D17-1155} {Instance weighting for
  neural machine translation domain adaptation}.
\newblock In \emph{Proceedings of the 2017 Conference on Empirical Methods in
  Natural Language Processing}, pages 1482--1488, Copenhagen, Denmark.
  Association for Computational Linguistics.

\bibitem[{Wattarujeekrit et~al.(2004)Wattarujeekrit, Shah, and
  Collier}]{wattarujeekrit2004pasbio}
Tuangthong Wattarujeekrit, Parantu~K. Shah, and Nigel Collier. 2004.
\newblock \href {https://doi.org/10.1186/1471-2105-5-155} {Pasbio:
  predicate-argument structures for event extraction in molecular biology}.
\newblock \emph{{BMC} Bioinform.}, 5:155.

\bibitem[{Wen et~al.(2013)Wen, Zheng, Jang, Xiang, and
  Penstein~Ros{\'e}}]{wen-etal-2013-extracting}
Miaomiao Wen, Zeyu Zheng, Hyeju Jang, Guang Xiang, and Carolyn
  Penstein~Ros{\'e}. 2013.
\newblock \href {https://www.aclweb.org/anthology/P13-2145} {Extracting events
  with informal temporal references in personal histories in online
  communities}.
\newblock In \emph{Proceedings of the 51st Annual Meeting of the Association
  for Computational Linguistics (Volume 2: Short Papers)}, pages 836--842,
  Sofia, Bulgaria. Association for Computational Linguistics.

\bibitem[{Ziser and Reichart(2017)}]{ziser-reichart-2017-neural}
Yftah Ziser and Roi Reichart. 2017.
\newblock \href {https://doi.org/10.18653/v1/K17-1040} {Neural structural
  correspondence learning for domain adaptation}.
\newblock In \emph{Proceedings of the 21st Conference on Computational Natural
  Language Learning ({C}o{NLL} 2017)}, pages 400--410, Vancouver, Canada.
  Association for Computational Linguistics.

\end{thebibliography}

\appendix
\section{Implementation Details}
\label{sec:supplemental}
\textbf{BERT}: The BERT baseline model uses the uncased variant of BERT-Base (with no additional fine-tuning) for feature extraction. We generate token representations by running BERT-Base and concatenating the outputs of the model's last 4 hidden layers. The BiLSTM layer has a hidden size of 100, with an input dropout of 0.5. The MLP layer is 100-dimensional. These values are consistent with the setup in \newcite{naik2020open}. \\
\textbf{BERT-ADA:} The domain predictor (adversary) is a 3-layer MLP with each layer having a dimensionality of 100 and ReLU activations between layers. For the hyperparameter $\lambda$, which is the constant used to weight domain prediction loss, we experiment with values from [0.5,1.0,2.0,5.0], and choose the best model based on F1 scores on the source domain validation set. We run one search trial with a fixed random seed (0) for all settings. The best performing model on clinical notes uses $\lambda=1.0$ and on conversations uses $\lambda=0.5$.\\
\textbf{BERT-LIW:} The autoregressive word-level language models used for weighting are 3-layer LSTMs, with a hidden size of 300 and layer dropout of 0.2 at each layer. Input embeddings are initialized using 300-dimensional GloVe embeddings, with parameter typing between input and output embedding matrices. The models are trained using SGD with gradient clipping at 0.25 and a batch size of 16 for 25 epochs. Training starts with a learning rate of 20, which is divided by 4 whenever validation loss plateaus.\\
\textbf{BERT-DAFT/BERT-DAFT-SYN:} BERT-Base is fine-tuned for 3 epochs, using a batch size of 4 and default parameter settings in the Huggingface transformers library. \\
All event extraction models are trained with a batch size of 16 and use Adam optimizer with a learning rate of 0.001. Models are trained for 1000 epochs with early stopping. All experiments are run on an NVIDIA GeForce GTX 1080 Ti machine.

\end{document}